# EVA²: Exploiting Temporal Redundancy in Live Computer Vision

Mark Buckler
*Cornell University*
mab598@cornell.edu

Philip Bedoukian
*Cornell University*
pbb59@cornell.edu

Suren Jayasuriya
*Arizona State University*
sjayasur@asu.edu

Adrian Sampson
*Cornell University*
asampson@cs.cornell.edu

*Abstract*—Hardware support for deep convolutional neural networks (CNNs) is critical to advanced computer vision in mobile and embedded devices. Current designs, however, accelerate generic CNNs; they do not exploit the unique characteristics of real-time vision. We propose to use the temporal redundancy in natural video to avoid unnecessary computation on most frames. A new algorithm, *activation motion compensation*, detects changes in the visual input and incrementally updates a previously-computed activation. The technique takes inspiration from video compression and applies well-known motion estimation techniques to adapt to visual changes. We use an adaptive key frame rate to control the trade-off between efficiency and vision quality as the input changes. We implement the technique in hardware as an extension to state-of-the-art CNN accelerator designs. The new unit reduces the average energy per frame by 54%, 62%, and 87% for three CNNs with less than 1% loss in vision accuracy.

*Keywords*-convolutional neural networks; computer vision; video compression; application specific integrated circuits; computer architecture, hardware acceleration;

## I. INTRODUCTION

Deep convolutional neural networks (CNNs) have revolutionized computer vision. A commensurate flurry of architecture work has designed efficient hardware CNN accelerators [1–4] targeting resource-strapped mobile and embedded systems. These designs, however, target generic convolutional networks: they do not exploit any domain-specific characteristics of embedded vision to improve CNN execution. This paper specializes CNN hardware for real-time vision on live video, enabling efficiency benefits that are unavailable to generic accelerators.

The key insight is that live vision is *temporally redundant*. In an input video, every frame differs only slightly from previous frames. A generic CNN accelerator, however, runs nearly identical, equally expensive computations for every frame. This traditional strategy wastes work to compute similar outputs for visually similar inputs. Existing work has also shown that CNNs are *naturally approximate*. Sparse accelerators, for example, improve efficiency with negligible impact on CNN output by rounding small network weights down to zero [3–6].

To exploit both temporal redundancy and approximation in real-time vision, we present an *approximately incremental* strategy for CNN execution. This strategy builds on ideas in incremental computation [7, 8]. To process an initial input frame, the strategy runs the full CNN and records both the input pixels and the output activation data. On subsequent frames, it detects changes in the pixels with respect to the saved input and uses these changes to update the saved output. The incremental update is much cheaper than full CNN execution, so the performance and energy benefits outweigh the cost of detecting input changes. The updated output, however, need not match the full CNN execution exactly; instead, the incremental update is a close approximation of the original CNN layers. We apply ideas from approximate computing [9–11] to adaptively control the trade-off between vision accuracy and resource efficiency by reverting to full computation when the prediction error is likely to be high.

Our algorithm, *activation motion compensation* (AMC), takes inspiration from video codecs [12] and literature on exploiting optical flow for computer vision [13, 14]. AMC captures visual motion in the input video and uses it to transform saved CNN activations. When pixels move in the input scene, AMC moves the corresponding values in the activation data. The algorithm skips a series of layers in the CNN by predicting their output and then invokes the remaining layers to compute the final vision result. We describe the mathematical relationship between pixel motion and convolutional layers and develop a new, hardware-optimized motion estimation algorithm to exploit this relationship.

We design a new hardware module, the *Embedded Vision Accelerator Accelerator* (EVA²*), that implements the AMC algorithm. Instead of designing a CNN accelerator from scratch, we show how to apply EVA² to any generic CNN accelerator to improve its efficiency for live computer vision. EVA² adds new logic and memories to a baseline accelerator to skip the majority of CNN layer executions for the majority of frames. EVA² uses an adaptive control scheme to decide which frames to run precisely.

We implement EVA² and synthesize the design on a 65 nm process. We augment state-of-the-art designs for accelerating convolutional and fully-connected layers [2, 6] and find that

---

*Pronounced *ee-vah squared*.

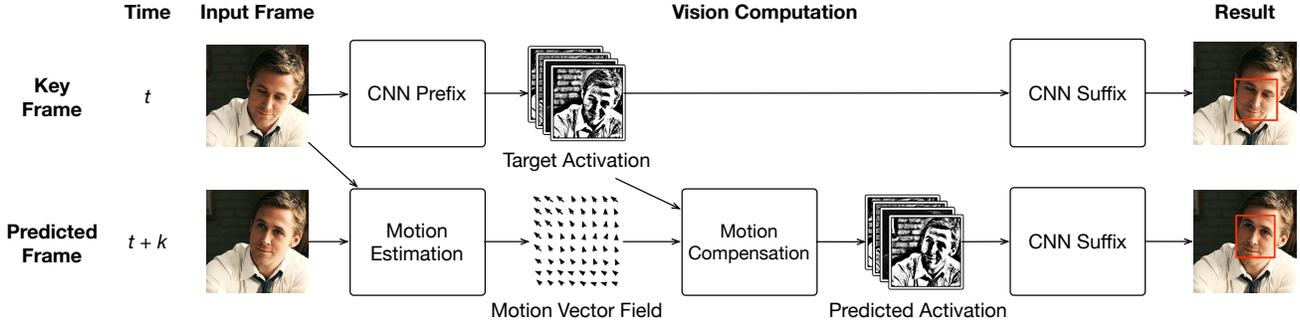

Figure 1: Activation motion compensation (AMC) runs the CNN precisely for periodic *key frames* and uses an approximately incremental update for more frequent *predicted frames*. For predicted frames, the algorithm estimates motion in the input and uses the resulting vector field to update a saved CNN activation from the last key frame.

EVA$^2$ makes up 3.5% of a full ASIC's area. On three CNN-based vision workloads, using EVA$^2$ reduces the average energy cost per frame by 54%, 62%, and 87% and decreases the average frame latency by similar amounts with less than 1% loss in vision task accuracy.

## II. ACTIVATION MOTION COMPENSATION

This section introduces activation motion compensation (AMC), our strategy for exploiting temporal redundancy in CNNs for efficient real-time computer vision. The central idea is to use saved outputs from earlier *key frames* to predict the output for later *predicted frames*. As in traditional incremental computing [7, 8], the challenge is finding an algorithm for updating saved results that is much cheaper than running the entire computation "from scratch" on the new input. Unlike traditional strategies, however, our approach for vision is *approximately incremental:* the updated output need not match the original output values precisely. Instead, predicted frames need only yield equivalently high vision accuracy results.

One natural strategy for approximately incremental CNN execution would exploit the differentiability of neural network layers. If $f(x)$ is a layer, then there must be another function $df$ such that $f(x+dx) \approx f(x) + df(dx)$. *Delta networks* operate by storing the old activation, $f(x)$, for every layer, computing $df(dx)$ for new layers, and adding it to the stored data [15, 16]. While delta updates are straightforward to implement, they do not address the primary efficiency bottlenecks in CNN execution. First, the hardware must store the activation data for every network layer to apply per-layer deltas, significantly increasing the memory requirements for CNN execution. Second, to compute $df(dx)$ for every layer, the hardware must load the full set of model weights, and previous work shows that the cost of loading weights can dominate the computation cost in CNNs [2, 17]. Finally, using pixelwise derivatives to represent changes in video frames assumes that individual pixels change their color slowly over time. If the camera pans or objects move in the scene, however, most pixels will change abruptly.

Instead of relying on a pixel-level derivative, our technique uses visual motion in the input scene. The intuition is the same as in video compression: most frames are approximately equal to the previous frame with some blocks of pixels moved around. AMC detects pixel motion and compensates for it in the output of a *single* target layer in the CNN. AMC builds on recent computer vision work to *warp* the target CNN activation data based on motion information [13]. Unlike delta updating, AMC bypasses the computation and memory accesses for an entire sequence of CNN layers.

### A. AMC Overview

Figure 1 illustrates AMC's approximately incremental execution strategy for real-time vision using CNNs. AMC processes the input video as as a mixture of *key frames*, which undergo full and precise CNN execution, and *predicted frames*, which use cheaper, approximate execution. AMC saves intermediate results during key frames and incrementally updates them for predicted frames. Section II-C4 describes how to decide which frames should be key frames.

To apply the strategy to a new CNN architecture, the system splits the CNN's series of layers into two contiguous regions: a larger prefix that only executes for key frames, and a smaller suffix that executes on every frame. The final layer in the prefix is called the *target layer:* AMC saves the output from this layer during key frames and predicts its output during predicted frames. Intuitively, the convolutional and pooling layers in the prefix have a spatial relationship to the input, while the suffix contains fully-connected layers and other computations where scene motion can have unpredictable effects. The prefix typically performs more generic feature extraction, while the functionality in later layers varies more between vision tasks [18]. Section II-C5 describes how to choose the target layer in more detail.

During key frames, AMC stores the network's input and the target layer's output for reuse during predicted frames. While the input image is typically small, CNN activations can have high dimensionality and occupy multiple megabytes [3]. Section II-C2 describes how to exploit activation sparsity to efficiently store key frame outputs.

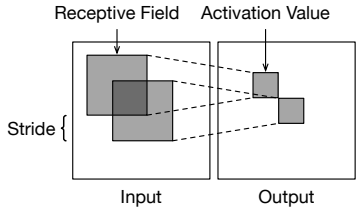

Figure 2: A convolutional layer applies filters to regions in the input. The input region corresponding to a given activation value is called its *receptive field*.

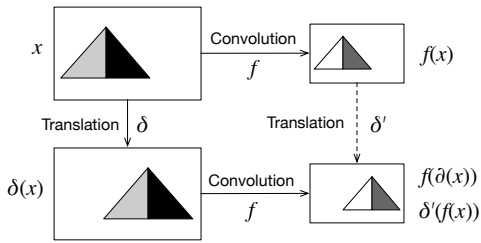

Figure 3: Convolutional layers and translations are *commutative*, so $f(\delta(x)) = \delta'(f(x))$ where $\delta'$ is a scaled version of the translation $\delta$.

During predicted frames, AMC detects the motion between the last stored key frame and the new input frame. *Motion estimation* is the problem of computing a vector field describing the visual displacement between two input frames. Motion estimation algorithms offer different trade-offs between granularity, accuracy, and efficiency [19–25]. AMC requires an efficient motion estimation technique that is aligned with the CNN's convolutional structure. Section II-C1 describes *receptive field block motion estimation* (RFBME), our new algorithm optimized for the hardware implementation of AMC.

Next, we describe mathematically how AMC updates stored activation data using motion information. Then, Section II-C describes the complete implementation of AMC.

### B. Warping CNN Activations

The core challenge in AMC is accurately updating saved activation data at the target layer according to input motion. *Activation warping*, as shown in Figure 1, takes in an old activation and a vector field and produces an updated activation. To perform warping, AMC needs to convert the vector field describing motion in the input image into a corresponding vector field for the target activation data. This conversion depends on the structure of convolutional layers, illustrated in Figure 2. Convolutional layers scan over the input using a fixed stride and compute the dot product of a filter matrix and a region in the input to produce each output value. This input region corresponding to each output value is called its *receptive field*. By propagating this structure through multiple convolutional layers, a receptive field in the input pixels can be defined for every activation value at every layer. Using receptive fields, we can derive the relationship between motion in the input image and in the target activation layer.

*Convolutions and translations commute.* The key insight is that, when the pixels in a receptive field move, they cause their corresponding activation value in the target layer to move by a similar amount. Mathematically, convolutional layers *commute* with translation: translating (i.e., moving) pixels in the input and then applying a convolution is the same as applying the convolution to the original pixels and then translating its output. (See Figure 3.)

Let $f$ be a convolutional layer in a CNN, let $x$ be an input image, and let $\delta$ be the vector field describing a set of translations in $x$. We define $\delta(x)$ to be the new image produced by translating pixels in $x$ according to the vector field $\delta$. We can also produce a new vector field, $\delta'$, by scaling $\delta$ by the stride of the convolution: for a convolutional layer with stride $s$, a distance $d$ in the input is equivalent to a distance $\frac{d}{s}$ in the output. The commutativity of translations and convolutions means that:

$$f(\delta(x)) = \delta'(f(x))$$

In AMC, $x$ is a key frame and $f(x)$ is the target activation for that frame. If $\delta(x)$ is a subsequent input frame, AMC can use the saved output $f(x)$ and compute $\delta'(f(x))$ instead of running the full computation $f$ on the new input.

Figure 4 shows an example convolution. If Figure 4a is a key frame, then Figure 4b shows a translation of the key frame to the right by 2 pixels. The output of that convolution is translated by the same amount in the same direction. The max-pooling output translates by only 1 pixel because pooling reduces the output resolution.

*Sources of approximation.* In this model, activation warping is perfectly precise: an incremental update produces the same result as the full CNN computation. However, the formulation relies on strong assumptions that do not generally hold for real video and complete CNNs. To the extent that these conditions are violated, AMC is approximate.

To clarify the sources of approximation in AMC, we list sufficient conditions under which AMC is precise. We demonstrate the conditions by example and describe how AMC mitigates—but does not eliminate—errors when the conditions are unmet. AMC relies on the inherent resilience in neural networks to tolerate the remaining imprecision.

*Condition 1: Perfect motion estimation.* AMC is precise when motion estimation perfectly captures the changes in the input. In other words, given a key frame $x_0$ and a predicted

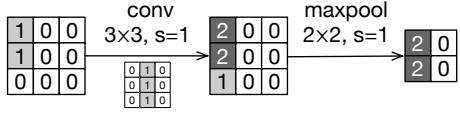

(a) An original image with a 3×3 convolutional layer and a 2×2 max-pooling layer applied, each with a stride (s) of 1. The inset shows the convolution's filter.

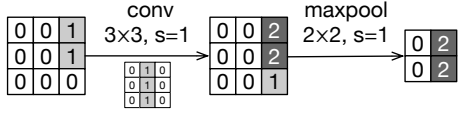

(b) The image in (a) translated to the right by 2 pixels. The outputs from the convolutional and pooling layers are similarly translated by 2 and 1 pixels respectively. Translation commutes precisely with the layers (accounting for the 2× loss in resolution from pooling).

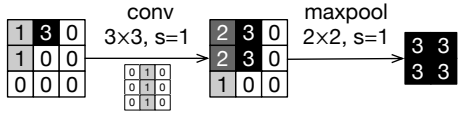

(c) The image in (a) with a "new pixel" that may have been revealed by de-occlusion. The new image is not a translation of the old image, so the outputs also have new values that cannot be reconstructed by translations.

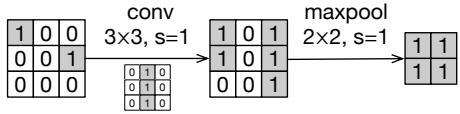

(d) The image in (a) with a single pixel translated. Because an entire 3×3 receptive field is not translated consistently, the output of the convolution is not a perfect translation of the original.

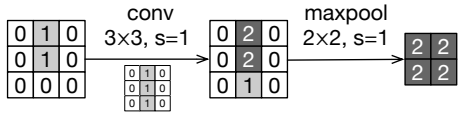

(e) The image in (a) translated to the right by 1 pixel. The output from the convolutional layer is translated, but the pooling layer produces a different output. Translation commutes precisely with the first layer but not the second.

Figure 4: An example convolutional layer and pooling layer applied to transformations of an input image.

frame $x_1$, motion estimation always finds a vector field $\delta$ such that $\delta(x_0) = x_1$ exactly. In reality, motion estimation algorithms are imperfect and not all changes between frames are attributable to motion.

For example, lighting changes and de-occlusion cause "new pixels" that are not present in the previous frame. Figure 4c shows an modified version of the input in Figure 4a with a new pixel. The output of the convolution, correspondingly, is not a translation of the original output.

*Condition 2: Convolution-aligned motion.* To perfectly translate the pixel vector field $\delta$ into an activation vector field $\delta'$, AMC assumes that blocks of pixels in receptive fields move in increments according to the convolution's stride. If a vector in $\delta$ has magnitude $d$, the corresponding vector in $\delta'$ has magnitude $\frac{d}{s}$ where $s$ is the convolutional layer's stride. If $d$ is not a multiple of $s$, the translation ends at a fractional coordinate in the activation and perfect reconstruction is impossible. Similarly, if pixels within a single receptive field move differently, a translation in the activation cannot perfectly capture the change.

Figure 4d shows a version of Figure 4a with a single pixel translated. Because the translation's granularity is smaller than a receptive field, the output of the convolution is not a perfect translation of its original output.

*Condition 3: Nonlinearities preserve motion.* Translation is commutative with convolution, but CNNs have other kinds of layers, such as pooling, that are not perfectly commutative. For example, Figures 4b and 4e show translated versions of the input in Figure 4a. In the first translation, a max-pooling layer produces an output that is a perfect translation of the original output. In the second, however, the max-pooling layer creates a non-translated output.

Real video and real CNNs violate all three of these conditions, so AMC's activation warping is an approximation of plain CNN execution. The central challenge in the design of the rest of the AMC algorithm is avoiding and suppressing these potential errors. AMC's adaptive key frame selection (Section II-C4) mitigates motion estimation errors by falling back to precise execution; interpolated warping (Section II-C3) addresses unaligned motion; and careful target layer selection (Section II-C5) avoids too much error accumulation due to nonlinear layers.

### C. Design Decisions for AMC

This section outlines our approaches to motion estimation, activation storage, warp interpolation, key frame selection, and target layer selection. AMC is a design space with a range of possible choices for each of these factors. We describe both the general requirements and the specific tactic we use in our hardware implementation.

*1) Efficient Motion Estimation.* The first step in executing an AMC predicted frame is motion estimation. Motion estimation takes two input frames and produces a 2D vector field describing the visual displacement between the frames. Motion estimation algorithms in the literature include block matching [19, 20], phase correlation in the Fourier domain [21], traditional optical flow algorithms such as Lucas–Kanade [22] and Horn–Schunck [23], and deep learning methods [24, 25]. A primary difference is the motion granularity: optical flow algorithms produce dense, pixel-level vector fields, while block matching detects coarser motion but can often be cheaper. Previous work on exploiting motion for efficient vision has used pixel-level optical flow [13]. For AMC, however, pixel-level motion estimation yields unnecessary detail: activation warping can only handle motion at the granularity of receptive fields. Another alternative is to use the motion vectors stored in compressed video data [26, 27], but real-time vision systems can save energy by skipping the ISP and video codec [28] to process uncompressed video streams.

Block matching algorithms, often used in video codecs, work by taking a block of pixels and comparing it to a window of nearby blocks in the reference (key) frame. The location of the closest matching reference block determines the motion vector for the block. Critical parameters include the choice of search window, the search organization, and the metric for comparing two candidate blocks [12].

We develop a new block matching algorithm, *receptive field block motion estimation* (RFBME), that is specialized for AMC. RFBME estimates the motion of entire receptive fields. The resulting displacement vector for each input receptive field maps to a corresponding displacement vector for a value in the target activation. RFBME avoids redundant work by dividing the input into square *tiles* whose size is equal to the receptive field stride and reusing tile-level differences. Section III-A describes the algorithm in detail and its hardware implementation.

*2) Compressed Activation Storage.* AMC needs to store the target activation data for the key frame so that subsequent predicted frames can reuse it. Activations for CNNs, however, can be large: storing an activation naively would require multiple megabytes of on-chip memory. To mitigate storage costs, we exploit *sparsity* in CNN activations. Many CNN accelerator proposals also exploit sparsity: most values in CNN weights and activations are close to zero, so they can be safely ignored without a significant impact on output accuracy [2–4, 6]. We use the same property to avoid storing near-zero values. Section III-B describes our hardware design for encoding and decoding sparse data.

*3) Interpolated Warping.* AMC's activation warping step takes a vector field $\delta$ and an old CNN activation and updates the activation according to the translations in the vector field $\delta'$. It works by scaling the magnitudes of the vector field to match the dimensions of the activation data. This scaling can produce vectors that are unaligned to activation coordinates. To translate by a fractional distance, AMC needs to interpolate the values of nearby activations.

There are a range of possible strategies for interpolation, from cheap techniques, such as nearest neighbor and bilinear interpolation, to more computationally expensive but accurate interpolation methods that preserve edge or gradient boundaries. For this paper, we choose bilinear interpolation to average neighboring pixels in 2D space while maintaining high performance. In our experiments, bilinear interpolation improves vision accuracy by 1–2% over nearest-neighbor matching on average for one CNN benchmark (FasterM).

*4) Selecting Key Frames.* The primary control that AMC has over vision accuracy and execution efficiency is the allocation of key frames, which are both more expensive and more accurate than predicted frames. Several strategies exist to decide when to use each type of frame. The simplest is a static key frame rate: every $n$th frame is a key frame, and the rest are predicted frames. A adaptive strategy, however, can allocate more key frames when the scene is chaotic and unpredictable and fewer when AMC's predictions are more likely to succeed. To implement an adaptive strategy, the accelerator must measure some feature of the input scene that correlates with the probability of a successful AMC prediction. We consider two possible features:

*Pixel compensation error.* AMC can produce a poor prediction when the motion estimation fails to accurately reflect the changes in the scene. To measure motion accuracy, we can reuse the internal bookkeeping of a block matching algorithm, which must compute the match error for each pixel block in the scene. When the aggregate error across all blocks is high, this strategy allocates a new key frame. When large occlusions occur, for example, this method will identify the inadequacy of the motion information.

*Total motion magnitude.* AMC's predictions are more accurate when there is less motion in the input scene. To measure the amount of motion, this simple strategy sums the magnitude of the vectors produced by motion estimation. This policy uses a key frame when the total amount of motion is large.

We implement and measure both techniques to compare their effectiveness. Section IV-E5 quantitatively compares their effects on overall accuracy.

*5) Choosing the Target Layer.* To apply AMC to a given CNN, the system needs to choose a target layer. This choice controls both AMC's potential efficiency benefits and its error rate. A later target layer lets AMC skip more computation during predicted frames, but a larger CNN prefix can also compound the influence of layers that make activation warping imperfect. Some kinds of layers, including fully-connected layers, make activation warping impossible: they have no 2D spatial structure and no meaningful relationship with motion in the input. These *non-spatial* layers must remain in the CNN suffix, after the target layer. Fortunately, these non-spatial layers are typically located later in CNNs, in the more task-specific segment of the network [18]. AMC also cannot predict through stateful structures in recurrent neural networks and LSTMs [29]. However, even these specialized networks tend to contain early sets of stateless convolutional layers for feature extraction [30, 31], where AMC can apply.

As with key frame selection, the system can use a static or adaptive policy to select the target layer. In our evaluation, we measure the impact of choosing earlier and later CNN layers, up to the last spatial layer (Section IV-E3) and find that the accuracy difference is negligible. Therefore, we implement AMC by statically targeting the last spatial layer. Dynamic policies for choosing the target layer may be useful future work if more complex networks demonstrate meaningful differences in accuracy for different input frames.

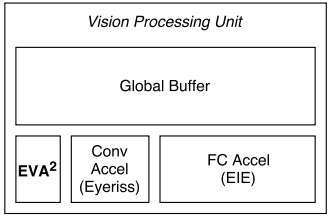

Figure 5: EVA$^2$ as a part of a complete vision processing unit (VPU). CNNs consist primarily of convolutional layers and fully-connected layers; this example VPU uses ASIC designs from the architecture literature for each of the two layer types [2, 6] and adds EVA$^2$.

## III. THE EMBEDDED VISION ACCELERATOR ACCELERATOR

This section describes the design of the Embedded Vision Accelerator Accelerator (EVA$^2$), our efficient hardware implementation of activation motion compensation. EVA$^2$ is not a complete CNN accelerator. Instead, we design it to complement existing deep learning accelerators. Figure 5 shows how EVA$^2$ fits into a complete vision processing unit (VPU) that also includes hardware for executing the convolutional and fully-connected layers that make up the bulk of CNN computation.

When the VPU processes a new frame, EVA$^2$ performs motion estimation and decides whether to run the computation as a key frame or a predicted frame. For key frames, EVA$^2$ sends the unmodified pixels to the layer accelerators and invokes them to run the full CNN. For predicted frames, EVA$^2$ instead performs activation warping and invokes the layer accelerators to compute the CNN suffix.

Figure 6 shows the high-level architecture of EVA$^2$. Two *pixel buffers* store video frames: one pixel buffer holds the most recent key frame and the other holds the current input frame. The *diff tile producer* and *diff tile consumer* cooperate to run motion estimation. The *key frame choice* module uses absolute pixel differences from the diff tile consumer to decide whether to treat it as a key frame or a predicted frame. For predicted frames, the diff tile consumer also sends a motion vector field to the *warp engine*, which updates the buffered key frame activation and sends the result to the layer accelerators. For key frames, the layer choice module toggles the muxing for the pixel buffers to reverse their roles and sends the pixels to the layer accelerators. When the layer accelerators send the target activation data back, EVA$^2$ stores it in its sparse key frame activation buffer.

EVA$^2$ makes no assumptions about layer computation, so only minor changes are required for any given CNN accelerator: the layer accelerators need additional muxing to receive activation inputs from EVA$^2$ during predicted frames, and the composite design needs to convert between the accelerator's native activation encoding and EVA$^2$'s run-length activation encoding.

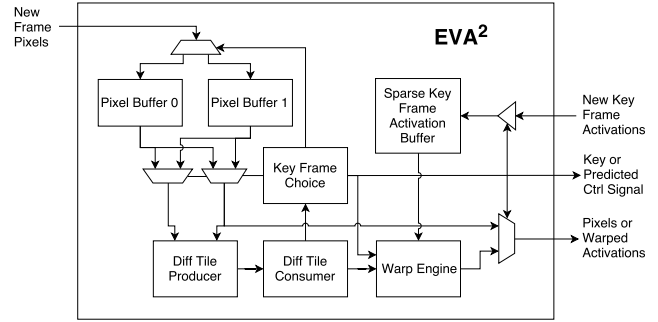

Figure 6: The architecture of EVA$^2$.

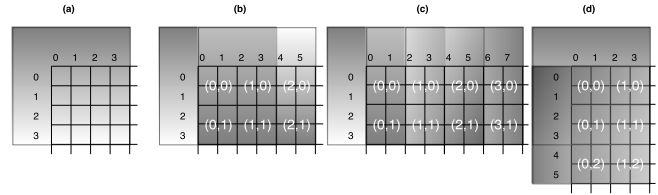

Figure 7: A set of example receptive fields with size 6, stride 2, and padding 2. Nearby receptive fields overlap significantly, and receptive fields near the edge overlap enclose out-of-bounds pixels.

The rest of this section describes the design of EVA$^2$'s two main logic components: the motion estimation logic, consisting of a diff tile producer and consumer, and the motion compensation logic, consisting of the warp engine.

### A. Receptive Field Block Motion Estimation

EVA$^2$ implements receptive field block motion estimation (RFBME), the specialized block matching algorithm motivated in Section II-C1. The algorithm exploits two properties of receptive fields for efficiency. First, the receptive field size is typically much larger than its stride, so nearby receptive fields overlap significantly and can reuse computation. Second, layer padding means that receptive fields often exceed the bounds of the image, and comparisons with these out-of-bounds pixels are unnecessary.

Figure 7 shows an example input matrix where the receptive fields have size 6, stride 2, and padding 2. The first receptive field (a) extends beyond the image bounds, so only 16 of its 36 pixels are valid. The second receptive field (b) entirely encloses the first receptive field's valid pixels, and the third (c) overlaps partially, in the range $x \in [2,3], y \in [0,3]$. The receptive fields in this example overlap on $2 \times 2$ *tiles*. In general, these tiles are $s \times s$ squares where $s$ is the receptive field's stride. RFBME divides the image and the receptive fields into tiles for comparison. When receptive field size is not an integer multiple of the stride, RFBME ignores partial tiles; we find these border pixels do not significantly impact RFBME's motion vectors.

The key insight is that the total absolute pixel difference for a receptive field is the sum of the differences of its tiles, and these tile differences are shared between many receptive fields. For example, imagine that the algorithm has already computed tile-level differences for all four tiles in the first receptive field, shown in Figure 7a, at a given comparison offset. The second receptive field (b) shares these same four tiles and adds two more: the tiles labeled $(2,0)$ and $(2,1)$. To compute a receptive field difference at the same offset, the algorithm can reuse the previously-computed differences for the first four tiles and add on differences for the two new tiles. The potential benefit from this reuse depends linearly on the number of pixels per tile. While the stride in this example is small, the stride in later layers of modern CNNs can be 16 or 32 pixels, which exponentially increases the amount of shared pixels per tile.

To exploit this reuse of tile differences, our RFBME microarchitecture uses a producer–consumer strategy. The *diff tile producer* compares $s \times s$ tiles to produce tile-level differences, and the *diff tile consumer* aggregates these differences to compute receptive field differences. The consumer then finds the minimum difference for each receptive field; this determines its offset and the output of RFBME. The next two sections describe each stage in detail.

*1) Diff Tile Producer.* For each tile, the diff tile producer performs a search across the key frame according to a fixed *search stride* and *search radius*. It uses a subsampled traditional exhaustive block matching search [12]. This search considers all locations in the key frame that are aligned with the search stride and are within the search radius from the tile's origin. A wider radius and a smaller stride yield higher accuracy at the expense of more computation.

To perform the search, the diff tile producer first loads a tile from the pixel buffer that contains the new input frame. Then, it iterates over the valid (in-bounds) search offsets in the old key frame according to the search parameters. For each offset, the producer computes an absolute pixel difference with the current tile using an adder tree. When the search is complete, the producer moves on to the next tile and starts the search again.

*2) Diff Tile Consumer.* The diff tile consumer receives tile differences from the producer and coalesces them into full receptive field differences. Figure 8 shows its architecture, including memories for caching reused tiles and partial sums and the pipelined adder trees for incremental updates. At a high level, the consumer slides a receptive-field-sized window across the frame and sums the tile differences within the window. It adds new tiles at the sliding window's leading edge and subtracts old tiles from its trailing edge.

The consumer receives tile differences, streaming in row by row, from the producer and stores them in a *tile memory*. It buffers the incoming differences until it receives all the tiles for a receptive field. For example, in Figure 7a, the consumer calculates the first receptive field difference after

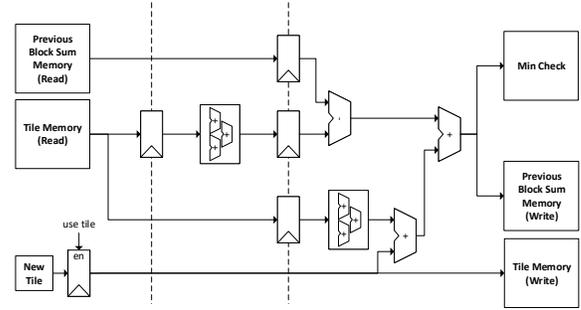

Figure 8: The architecture of the diff tile consumer used for receptive field block motion estimation (RFBME).

it receives the tile difference for $(1,1)$ in the second row.

To enable computation reuse, the consumer stores each new receptive field difference it computes in a *past-sum* memory. To compute a new receptive field difference, it loads the difference for the previous overlapping receptive field, adds a new column tile differences, and subtracts the old column of tile differences. For example, in Figure 7c, the second receptive field difference is the sum of the differences for the tiles $(0,0)$ through $(2,1)$. To calculate the difference for the third receptive field, the consumer fetches this value from the past-sum memory, adds the new tile column $(3,0)$ and $(3,1)$, and subtracts the old tile column $(0,0)$ and $(0,1)$. These rolling additions and subtractions avoid the need for exhaustive sums in the steady state.

The consumer checks every new receptive field difference against a single-entry *min-check* register to find the minimum difference. When it finds a new minimum value, the consumer writes the difference and offset back to the min-check memory.

When the consumer finally finishes processing all the receptive fields, it sends the minimum-difference offsets, which constitute motion vectors, to the warp engine. It also sends the minimum differences themselves to the key frame choice module, which uses the total to assess the aggregate quality of the block matching.

### B. Warp Engine

The warp engine, shown in Figure 9, uses motion vectors to update the stored activation data. Each motion vector ends at a fractional destination between four neighboring activation values. The warp engine's job is to load this neighborhood of activation values from its sparse activation memory, feed them into a bilinear interpolator along with the fractional bits of this motion vector, and send the result to the layer accelerators to compute the CNN suffix.

The first step is to load the activation data. EVA$^2$ uses run-length encoding (RLE) for activations. RLE is critical to enabling on-chip activation storage: for Faster16, for example, sparse storage reduces memory requirements by more than 80%. However, the sparse data representation complicates the task of loading individual values.

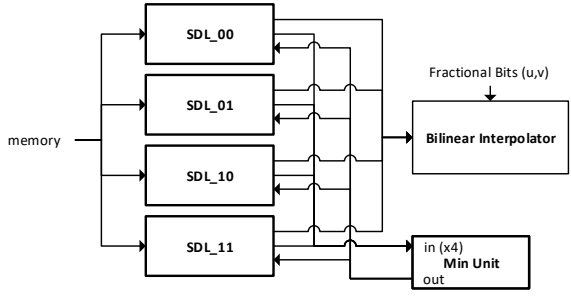

Figure 9: The architecture of the warp engine.

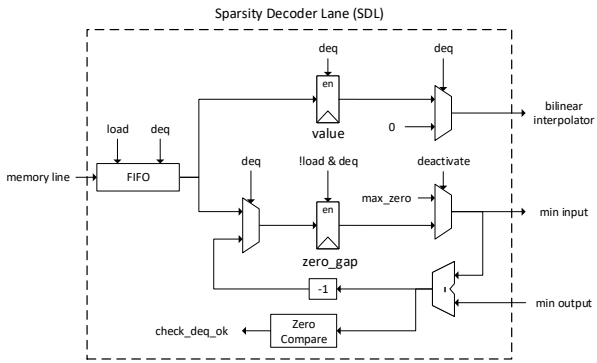

Figure 10: The warp engine's datapath for loading sparse activation data from the key activation memory.

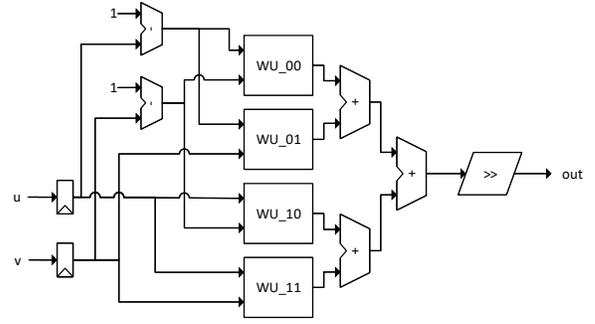

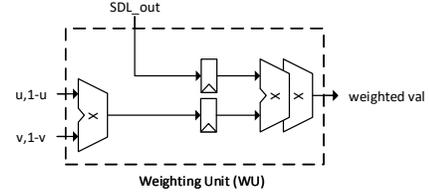

Figure 11: The warp engine's bilinear interpolation logic.

To load sets of four activation values, the warp engine uses four *sparsity decoder lanes* (Figure 10) to skip zeros shared among all four activations. Once activations are loaded into the FIFO, each lane sends its zero gap for its first activation. The min unit checks the zero gap in each lane and sends the minimum to all of the lanes. Each lane then decrements its zero gap by this amount, thereby jumping forward in the channel. All lanes with zero gaps of zero after the min subtraction provide their value register as input to the bilinear interpolator. Lanes with positive zero gaps provide zero as the input to the interpolator.

The warp engine feeds activation outputs from the sparsity decoder lanes into the four weighting units in the bilinear interpolator (shown in Figure 11). The interpolator is a two-stage datapath that computes a 4-way weighted sum using the activation values from the sparsity decoder lanes, SDL_00 through SDL_11, and the fractional bits of a given motion vector, $(u,v)$. It computes the weighted sum:

$$SDL\_00 \cdot (1-u) \cdot (1-v) + SDL\_01 \cdot (1-u) \cdot v \\ + SDL\_10 \cdot u \cdot (1-v) + SDL\_11 \cdot u \cdot v$$

The interpolator computes wide intermediate values and then shifts the final result back to a 16-bit fixed-point representation. The warp engine sends the output activations to the layer accelerators to begin the CNN suffix computation.

## IV. EVALUATION

This section studies EVA$^2$'s impacts on vision accuracy and efficiency and explores its implementation trade-offs. We begin with a simple first-order model to build intuition and then proceed to a full, empirical evaluation.

### A. First-Order Efficiency Comparison

As Section II describes, AMC relies on a fundamental efficiency trade-off: for predicted frames, it eliminates the CNN prefix computation in exchange for incurring the cost of motion estimation and compensation. The technique succeeds if the former is much more expensive than the latter. To provide intuition for this advantage, we build a first-order model for these computation costs.

The cost of computation in the CNN prefix is dominated by the multiply-and-accumulate operations (MACs) in the convolutional layers. Each layer computes a number of outputs dictated by its size and filter count (channels):

outputs = layer width $\times$ layer height $\times$ out channels

Each output is the weighted sum of inputs within an area given by the filters' width and height:

MACs per output = in channels $\times$ filter height $\times$ filter width

We sum the total number of MACs for all convolutional layers in the prefix:

$$\text{prefix MACs} = \sum_{i}^{\text{prefix layers}} \text{outputs}_i \times \text{MACs per output}_i$$

For a Faster16 prefix ending at layer conv5_3 on 1000×562 images, for example, the total is $1.7 \times 10^{11}$ MACs.

AMC's cost for predicted frames, in contrast, consists mainly of motion estimation and compensation. The compensation step is a simple linear-time interpolation, so motion estimation dominates. Our motion estimation algorithm, RFBME (Section III-A), consists primarily of the additions and subtractions that accumulate absolute pixel differences. We first analyze an *unoptimized* variant of RFBME that does not exploit tile-level computation reuse. The algorithm sweeps a receptive field over a search region in the input image with a given radius and stride. At each point, the algorithm takes the difference for every pixel in the receptive field's area, so the total number of operations is given by:

$$\text{unoptimized ops} = (\text{layer width} \times \text{layer height}) \times \left(\frac{2 \times \text{search radius}}{\text{search stride}}\right)^2 \times \text{rfield size}^2$$

The full RFBME algorithm reuses computations from *tiles* whose size is equal to the receptive field stride. It then incurs additional operations to combine the differences from tiles:

$$\text{RFBME ops} = \frac{\text{unoptimized ops}}{\text{rfield stride}^2} + (\text{layer width} \times \text{layer height}) \times \left(\frac{\text{rfield size}}{\text{rfield stride}}\right)^2$$

Again using Faster16 as an example, an unoptimized version requires $3 \times 10^9$ add operations while RFBME requires $1.3 \times 10^7$. Overall, for this example, AMC eliminates $\sim 10^{11}$ MACs in the CNN prefix and incurs only $\sim 10^7$ additions for motion estimation. AMC's advantages stem from this large difference between savings and overhead.

### B. Experimental Setup

We implement EVA$^2$ in RTL and synthesize our design using the Synopsys toolchain, including the Design Compiler, IC Compiler, and PrimeTime, in the TSMC 65 nm process technology. For energy and timing, we run simulations of the full EVA$^2$ design. The EVA$^2$ implementation uses eDRAM memories for the three larger buffers: the two pixel buffers and the activation buffer. We use CACTI 6.5 [32] to measure the memories' power, performance, and area. CACTI includes both an eDRAM model and an SRAM model, which we use for EVA$^2$'s smaller buffers.

*Baseline accelerator.* We apply EVA$^2$ to a model of a state-of-the-art deep learning accelerator based on recent architecture papers. We model Eyeriss [2] for convolutional layers and EIE [6] for fully-connected layers by gathering published per-layer results from each paper. A journal paper about Eyeriss [33] reports both power and latency numbers for each layer in the VGG-16 [34] and AlexNet [35] networks. The EIE results only include latency numbers for these two networks, so we use the total design power to estimate energy. Because EIE is implemented on a TSMC 45 nm process, we normalize by scaling up the power, latency, and area for EIE according to the technology scaling factor. Eyeriss is implemented in the same TSMC 65 nm technology and thus requires no scaling. The total latency and energy for a CNN execution in our model is the sum of the individual layer costs. To quantify the cost of layers not present in AlexNet and VGG-16, the model scales the average layer costs based on the number of multiply–accumulate operations required for each layer, which we find to correlate closely with cost in both accelerators [36].

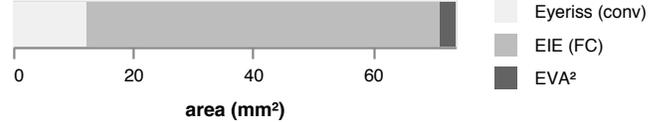

Figure 12: Hardware area on a 65 nm process for EVA$^2$ compared to deep learning ASICs: Eyeriss [2] for convolutional layers and EIE [6] for fully-connected layers.

*EVA$^2$ hardware.* Our RTL implementation of EVA$^2$ meets timing with a clock cycle of 7 ns, which was matched to the memory cycle time. Figure 12 compares the area for EVA$^2$ against the reported area for the convolutional layer accelerator (Eyeriss) and the fully-connected accelerator (EIE). The area for Eyeriss [2] is 12.2 mm$^2$ on a 65 nm process, 78.6% of which is occupied by its PEs. The area for EIE [6] is 40.8 mm$^2$ on a 45 nm process; compensating for the process difference, EIE would occupy approximately 58.9 mm$^2$ on a 65 nm process. EVA$^2$ itself occupies 2.6 mm$^2$, which is 3.5% of the overall area for the three units. Of this, the eDRAM memory for the pixel buffers occupies 54.5% of EVA$^2$'s area, and the activation buffer occupies 16.0%.

*Vision applications and dataset.* We study three convolutional neural networks. **AlexNet** [35] is an object classification CNN consisting of 5 convolutional layers and 3 fully-connected layers. **Faster16** is an version of the Faster R-CNN object detection network [37], which can use different networks for its feature extraction phase, based on the VGG-16 recognition network [34]. VGG-16 has 16 convolutional layers; Faster R-CNN adds 3 convolutional layers and 4 fully-connected layers. **FasterM** is a different variant of Faster R-CNN based on the "medium" CNN-M design from Chatfield et al. [38]. CNN-M has only 5 convolutional layers, so it is smaller and faster but less accurate than VGG-16. We use standard vision metrics to assess EVA$^2$'s impact on accuracy. For our classification network, AlexNet, we use top-1 accuracy; for the object detection networks, we use

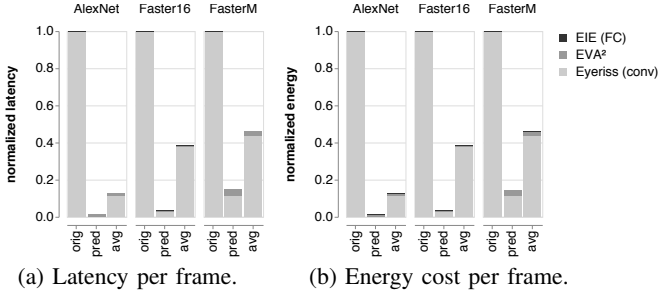

(a) Latency per frame.  (b) Energy cost per frame.

Figure 13: Performance (a) and energy (b) impact of $EVA^2$. *Orig* shows the baseline CNN execution, *pred* shows the cost of predicted frames with $EVA^2$, and *avg* shows the overall average cost per frame.

| Network | Config | Acc. | Keys | Time (ms) | Energy (mJ) |
|---|---|---|---|---|---|
| AlexNet | orig | 65.1 | 100% | 115.4 | 32.2 |
|  | hi | 65.1 | 22% | 26.7 | 7.4 |
|  | med | 64.3 | 11% | 14.5 | 4.0 |
|  | lo | 63.8 | 4% | 5.9 | 1.6 |
| Faster16 | orig | 60.1 | 100% | 4370.1 | 1035.5 |
|  | hi | 60.0 | 60% | 2664.8 | 631.3 |
|  | med | 59.4 | 36% | 1673.6 | 396.4 |
|  | lo | 58.9 | 29% | 1352.7 | 320.3 |
| FasterM | orig | 51.9 | 100% | 492.3 | 116.7 |
|  | hi | 51.6 | 61% | 327.2 | 77.4 |
|  | med | 51.3 | 37% | 226.4 | 53.4 |
|  | lo | 50.4 | 29% | 194.7 | 45.9 |

Table I: The trade-off space between accuracy and resource efficiency with $EVA^2$. For the original baseline and three key configurations, we show the vision task accuracy score (*acc*), the fraction of frames that are key frames (*keys*), and the average latency and energy cost per frame.

mean average precision (mAP).

To train, validate, and test the CNNs, we use the Google YouTube-BoundingBoxes dataset (YTBB) [39]. This dataset consists of 240,000 videos annotated with object locations and categories. It includes ground truth for both object detection and frame classification. Training, testing, and reference frames were all decoded at 30 frames per second, corresponding to a 33 ms time gap between each frame. The total dataset is more than five times larger than ImageNet [40], so we use subsets to reduce the time for training and testing. We train on the first 1/25 of the training datasets (132,564 and 273,121 frames for detection and classification, respectively). To evaluate during development, we used a validation dataset consisting of the first 1/25 of each of YTBB's validation sets (17,849 and 34,024 frames for the two tasks). Final results reported in this section use a fresh test set consisting of the last 1/25 of YTBB's validation sets.

We use hyperparameters without modification from open-source Caffe [41] implementations of each network. All three networks were initialized with weights pretrained on ImageNet [40]. We train on a Xeon server using two NVIDIA GTX Titan X Pascal GPUs.

### C. Energy and Performance

Figure 13 depicts the energy savings that $EVA^2$ offers over the baseline CNN accelerator. The figure shows the energy and latency cost for processing a single frame on the baseline accelerator without $EVA^2$, the average with $EVA^2$ enabled, and the costs for $EVA^2$'s predicted frames alone. In these configurations, the degradation of the application's vision quality score is at most 1 percentage point. Even the smallest savings are significant. For FasterM, the energy cost with $EVA^2$ is 46% of the baseline cost. The savings are particularly dramatic for AlexNet because $EVA^2$ adapts to an extremely low key frame rate for classification; the next section describes this effect in more detail.

The energy and latency for the fully-connected layers are orders of magnitude smaller than for convolutional layers. This difference is due to EIE's efficiency: it exploits the redundancy in fully-connected layer weights to store the entire model on chip [6]. This on-chip storage is reflected in EIE's low latency and energy and its large area requirement (see Figure 12). Eyeriss's use of off-chip DRAM is representative of other CNN ASIC designs [3].

### D. Accuracy–Efficiency Trade-Offs

To quantify AMC's trade-off space between efficiency and accuracy, we consider three configurations with different key frame rates. Table I lists the accuracy and efficiency of three configurations, *hi*, *med*, and *lo*, found by limiting the task accuracy degradation on the validation set to $<0.5\%$, $<1\%$, and $<2\%$, respectively. We measure the average frame cost and accuracy on a test set. The *med* configuration is also shown in Figure 13.

The measured accuracy drop in every configuration is small, and $EVA^2$'s benefits improve as the accuracy constraint is loosened. For FasterM's *lo* configuration, for example, only 29% of the frames are key frames, so the average energy cost per frame is only 39% of the baseline, but the test-set accuracy drop remains less than 1.5%.

For AlexNet, extremely low key frame rates suffice. Even in the *hi* configuration, which has no accuracy drop within three significant figures, only 22% of frames are key frames. The reason is that, unlike object detection results, frame classification results change slowly over time. $EVA^2$'s adaptive key frame selection can help decide when a new class may be likely, but the vast majority of frames have the same class as the previous frame.

### E. Design Choices

We characterize the impact of various design choices in $EVA^2$'s implementation of AMC.

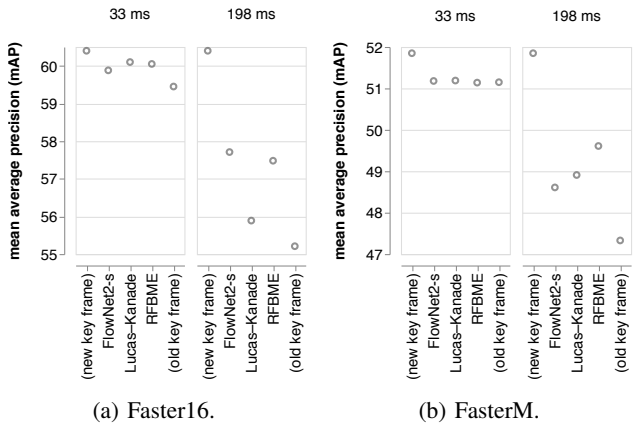

(a) Faster16.  (b) FasterM.

Figure 14: Accuracy impact of motion estimation techniques. *New key frame* shows the ideal baseline accuracy when computing the full CNN precisely, *old key frame* shows the worst-case accuracy for using the previous key frame without updating it at all, and the rest are motion estimation algorithms. RFBME is our new algorithm.

*1) Motion Compensation vs. Memoization.* The first choice when using AMC is whether it should use motion compensation to update the key frame's target activation or just reuse the previous result without modification (i.e., simple memoization). The choice depends on the vision task. Some applications, such as object detection, semantic segmentation, and pose estimation are very sensitive to pixel translation. For these applications, motion compensation improves accuracy over memoization. This includes Faster16 and FasterM, which are object detection networks: Figure 14 illustrates motion compensation's benefit over using the old key frame. Other tasks, such as classification, are designed to be insensitive to translation. For networks like AlexNet, motion compensation does not improve its predictions and can even degrade them by introducing noise. The accuracy degradation for AlexNet with a key frame gap of 4891 ms under simple memoization is only 1%, but enabling motion compensation worsens the change to 5%. As a result, we use memoization for AlexNet and full motion compensation for FasterM and Faster16 in the rest of this evaluation.

*2) Motion Estimation.* Motion estimation is a key ingredient for AMC. We compare our custom block-based motion estimation algorithm, RFBME, with two pixel-level optical flow techniques: the classic Lucas–Kanade algorithm [22] and the FlowNet 2.0 CNN [25]. Unlike RFBME, both algorithms produce pixel-level vector fields. To convert these to receptive-field-level fields, we take the average vector within each receptive field.

Figure 14 shows the overall mean average precision for predicted frames in FasterM and Faster16 when using each method. We show the error when predicting at two time intervals from the key frame, 33 ms and 198 ms. At 33 ms, the predicted frame directly follows the key frame, so the amount of motion is small. We choose 198 ms because it consistently reveals accuracy degradation in the detection benchmarks. In each case, RFBME yields either the best or nearly the best accuracy. Because its efficiency does not come at a disadvantage in vision accuracy, we choose RFBME for our final EVA$^2$ design.

| Network | Interval | Early Target | Late Target |
|---|---|---|---|
| AlexNet | orig | 63.52 | 63.52 |
|  | 4891 ms | 49.95 | 53.64 |
| Faster16 | orig | 60.4 | 60.4 |
|  | 33 ms | 60.29 | 60.05 |
|  | 198 ms | 55.44 | 57.48 |
| FasterM | orig | 51.85 | 51.85 |
|  | 33 ms | 50.90 | 51.14 |
|  | 198 ms | 48.77 | 49.61 |

Table II: The accuracy impact of targeting different layers for EVA$^2$'s prediction at various key frame intervals. The *orig* rows show the baseline accuracy for each network.

| Network | Target Layer | Accuracy |
|---|---|---|
| FasterM | No Retraining | 51.02 |
|  | Early Target | 45.35 |
|  | Late Target | 47.82 |
| Faster16 | No Retraining | 60.4 |
|  | Early Target | 61.30 |
|  | Late Target | 60.52 |

Table III: The accuracy impact of training to fine-tune CNNs for execution on warped data. The accuracy column shows the network's score when processing plain, unwarped activation data.

*3) Target Layer.* The choice of the *target layer* for AMC controls the amount of savings it can offer and its error. We study the accuracy impact of selecting an early target layer and a late target layer for each application. In each case, the early layer is after the CNN's first pooling layer, and the late layer is the last *spatial* layer: the layer before the first fully-connected layer or other computation would prevent activation warping. Table II shows the accuracy for AMC predicted frames when selecting each of these layers as the target layer. In most cases, the accuracy for predicting at the later layer is *higher* than for the earlier layer. The exception is Faster16 at 33 ms, where the difference is small. This improvement suggests that AMC's activation updates are accurate enough even for a large CNN prefix. We use the later layer for each application in the rest of this evaluation.

*4) Training on Warped Activation Data.* While EVA$^2$'s predictions approximate "true" activation data, it may introduce artifacts that interfere with the normal operation of the CNN suffix. To counteract these artifacts, we can retrain the CNN suffix on warped activation data. Table III examines the impact of this retraining on FasterM and

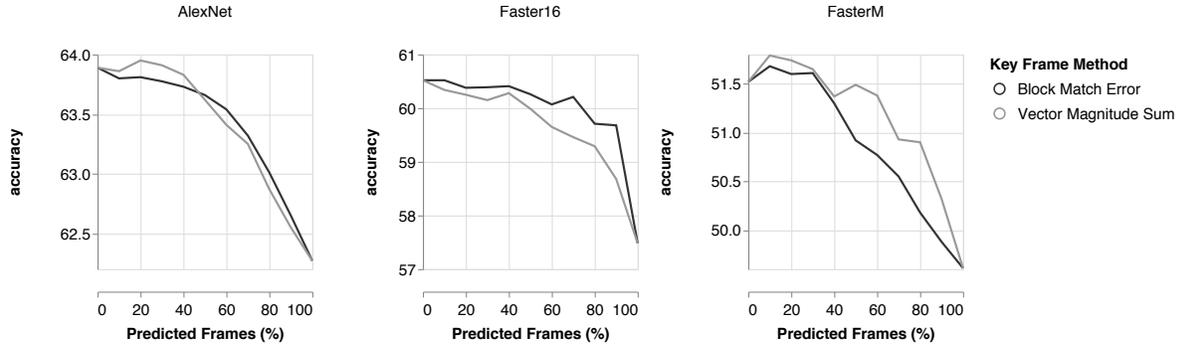

Figure 15: The impact of adaptive key frame selection strategy on vision accuracy. Each data point contains the same number of frames but with a different percentage of predicted frames. The *y*-axes show overall accuracy, and the time gap between key frames and predicted frames is fixed at 198 ms for Faster16 and FasterM, and 4891 ms for AlexNet.

Faster16 by measuring the resulting accuracy on plain (key) frames. For Faster16, the impact of retraining is small for the early target layer and negligible for the later target layer. For FasterM, retraining actually *decreases* accuracy on key frames, although this may be due to its limited training schedule in comparison with Faster16. We conclude that additional training on warped data is unnecessary.

*5) Key Frame Selection.* AMC can choose adaptively when to use expensive key frames and when to use cheap predicted frames. Section II-C4 describes two possible metrics that EVA$^2$ can use to choose key frames: RFBME match error and total motion magnitude. Figure 15 compares the vision accuracy for each strategy. In these experiments, we fix a key frame interval and sweep the decision threshold, then measure the resulting fraction of predicted frames and the output accuracy. For a fair comparison between the two metrics, each data point contains the same percentage of predicted frames and key frames for each metric. A fixed key frame rate would appear as a straight line from the 0% predicted frames point to the 100%. The curves for both metrics are above this fixed-rate line, so both are viable strategies. We use the block error metric in our hardware implementation because it is computationally cheap: block errors are byproducts of RFBME.

## V. RELATED WORK

EVA$^2$ builds on a wave of recent architectural work on efficient hardware for deep learning. For a survey of the state of the art, see the tutorial by Sze et al. [42]. Commercial hardware efforts include GPUs and manycores with customized vector and low-precision instructions [43, 44]. In research, many recent ASICs target convolutional and fully-connected layers [1, 2, 17, 45–48], as do some FPGA designs [49–52]. Recently, accelerators have focused on exploiting sparsity in model weights and activations [3, 4, 6, 53] and extreme quantization to ternary or binary values [54–56]. EVA$^2$'s benefits are orthogonal to these design choices. Because it skips entire layers during forward execution, it can apply to any underlying CNN accelerator architecture.

The AMC algorithm uses insights from video compression. Specifically, our RFBME algorithm builds on a large body of work on the block-matching motion estimation algorithms that are central to video codecs [19, 20, 57] and their ASIC implementations [58, 59].

In vision, the most closely related work is deep feature flow (DFF) [13, 14]. DFF is a neural network design that grafts an optical flow network, FlowNet [24], onto a subset of a feature network, ResNet [60], via a spatial warping layer. The goal is similar to AMC: DFF uses motion information to avoid computing a prefix of convolutional layers involved in feature extraction. AMC's focus on hardware efficiency provides four key benefits over DFF. (1) While RFBME performs coarse-grained computation at the receptive-field level, DFF uses a full CNN (FlowNet) to compute per-pixel motion, which is far more expensive. (2) DFF uses a fixed key frame rate calibrated to the "worst-case scenario" for scene motion. AMC's adaptive key frame rate spends less time and energy when frames are more predictable. (3) AMC's activation compression reduces the intermediate data size enough for on-chip storage (80–87%). (4) EVA$^2$'s warp engine skips over zero entries when performing interpolation, reducing the motion compensation cost proportionally to the activations' sparsity.

Other vision work has sought to exploit temporal redundancy for efficiency. Zhang et al. leverage motion vectors and residuals in compressed video to speed up super-resolution algorithms [26]. Future work may adapt AMC to replace RFBME with these precomputed motion vectors. Clockwork convnets [61] exploit the observation that the values in deeper, more semantic layers change at a slower rate than earlier, noisier layers. The execution strategy uses fixed update rates, however, and does not adapt to changes in the input video. Delta networks [15, 16] compute the temporal derivative of the input and propagate the change through a network, layer by layer. Using pixel-level derivatives, however, is a poor match for real video data, where even small movements can cause large changes in pixel values. Section II discusses delta networks and their efficiency drawbacks in more detail.

## VI. CONCLUSION

Generic CNN accelerators leave efficiency on the table when they run real-time computer vision workloads. While this paper exploits temporal redundancy to avoid CNN layer computation, AMC also suggests opportunities for savings in the broader system. Future work can integrate camera sensors that avoid spending energy to capture redundant data [28, 62–64], and end-to-end visual applications can inform the system about which semantic changes are relevant for their task. A change-oriented visual system could exploit the motion vectors that hardware video codecs already produce, as recent work has done for super-resolution [26]. Through holistic co-design, approximately incremental vision can enable systems that spend resources in proportion to relevant events in the environment.

ACKNOWLEDGMENTS

We thank the reviewers for their insightful comments, Khalid Al-Hawaj for his kind help with synthesis tools, and Skand Hurkat for reading a draft. This work was supported by a gift from Huawei. We also thank NVIDIA Corporation for donating a Titan X Pascal GPU used in this research.